\documentclass[10pt, a4paper]{article}
\usepackage{lrec2022} 
\usepackage{multibib}
\newcites{languageresource}{Language Resources}
\usepackage{graphicx}
\usepackage{tabularx}
\usepackage{soul}
\usepackage{titlesec}
\titleformat{\section}{\normalfont\large\bfseries\center}{\thesection.}{1em}{}
\titleformat{\subsection}{\normalfont\SmallTitleFont\bfseries\raggedright}{\thesubsection.}{1em}{}
\titleformat{\subsubsection}{\normalfont\normalsize\bfseries\raggedright}{\thesubsubsection.}{1em}{}
\renewcommand\thesection{\arabic{section}}
\renewcommand\thesubsection{\thesection.\arabic{subsection}}
\renewcommand\thesubsubsection{\thesubsection.\arabic{subsubsection}}
\usepackage[T1]{fontenc}

\usepackage{epstopdf}
\usepackage[utf8]{inputenc}

\usepackage{hyperref}
\usepackage{xstring}

\usepackage{color}
\usepackage{linguex}

\title{Domain Adaptation in Neural Machine Translation using a Qualia-Enriched FrameNet}

\name{Alexandre Diniz Costa, Mateus Coutinho Marim, Ely Edison da Silva Matos, \\
{\bf \large Tiago Timponi Torrent}\\}

\address{Federal University of Juiz de Fora | FrameNet Brasil \\
         Rua José Lourenço Kelmer, s/nº – Campus Universitário – Juiz de Fora, MG – Brazil \\
         \{alexandre.costa,ely.matos,tiago.torrent\}@ufjf.br, mateus.marim@ice.ufjf.br\\}

\abstract{
In this paper we present Scylla, a methodology for domain adaptation of Neural Machine Translation (NMT) systems that make use of a multilingual FrameNet enriched with qualia relations as an external knowledge base. Domain adaptation techniques used in NMT usually require fine-tuning and in-domain training data, which may pose difficulties for those working with lesser-resourced languages and may also lead to performance decay of the NMT system for out-of-domain sentences. Scylla does not require fine-tuning of the NMT model, avoiding the risk of model over-fitting and consequent decrease in performance for out-of-domain translations. Two versions of Scylla are presented: one using the source sentence as input, and another one using the target sentence. We evaluate Scylla in comparison to a state-of-the-art commercial NMT system in an experiment in which 50 sentences from the Sports domain are translated from Brazilian Portuguese to English. The two versions of Scylla significantly outperform the baseline commercial system in HTER.
\newline
\\ }

\begin{document}

\maketitleabstract

\section{Introduction}

Neural models have been advancing the state-of-art in the field of Machine Translation (MT) in a myriad of tasks \cite{barrault-etal-2019-findings,barrault-etal-2020-findings}. Nonetheless, as pointed out by \newcite{koehn-knowles-2017-six}, maintaining the high performance of Neural Machine Translation (NMT) in a specific domain for which there is a lack of large training data is challenging. Domain adaptation has been used as a strategy to mitigate such a loss in performance.

\newcite{DBLP:journals/corr/abs-1806-00258} pointed out that a big research question still to be answered is how to use external knowledge such as dictionaries and knowledge bases for domain adaptation in NMT. In this paper, we provide an answer to such a question in the form of Scylla, a methodology for domain adaptation of NMT systems. Scylla substitutes domain-specific terms in the source language by their adequate translation in the target language and is implemented in two versions: one of them, Scylla-S, does it before the source sentence is fed into the NMT system, while the other, Scylla-T, takes as input the already translated sentence. 

Both pipelines make use of a multilingual FrameNet covering the Sports domain \cite{Costa2017,Costa2018} as an external knowledge base. This is to say that they do not require any fine-tuning of the NMT system, avoiding model over-fitting and decrease in performance for general-domain translations \cite{khayrallah-etal-2018-regularized,thompson-etal-2019-overcoming}.

To evaluate Scylla, we conducted an experiment where 50 Brazilian Portuguese (br-pt) sentences, collected from Sports news and encyclopedias, were submitted to the commercial NMT system alone - considered as the baseline -, and to the two pipelines presented in this paper for translation into English (en). Systems' performances were then evaluated against BLEU \cite{papineni-etal-2002-bleu}, TER and HTER \cite{snover-etal-2006-study}. The two domain adaptation solutions presented in this paper significantly outperform the baseline for HTER.

The contributions of this paper are two-fold:
\begin{enumerate}
    \item We present two solutions for using a semantically structured external resource – FrameNet – for domain adaptation in NMT, both of which outperform the baseline;
    \item The solutions proposed do not require fine-tuning of the NMT model, substantially reducing computational costs.  
\end{enumerate}

In the remainder of this paper, we survey, in section \ref{sec:related}, recent research work focusing on the use of external resources for domain adaptation in MT. Section \ref{sec:fn-br} presents the FrameNet model used in the methodology. Scylla is presented in section \ref{sec:solutions} and its evaluation is discussed in section \ref{sec:evaluation}.

\section{ Related Work}
\label{sec:related}

In a survey paper on domain adaptation in NMT, \newcite{DBLP:journals/corr/abs-1806-00258} list three works using external knowledge for domain adaptation in MT: \newcite{arthur-etal-2016-incorporating}, \newcite{DBLP:journals/corr/ZhangZ16c} and \newcite{DBLP:journals/corr/abs-1709-02184}. In this section we provide a summary of each of them and also of \newcite{moussallem2019utilizing} and \newcite{dougal-lonsdale-2020-improving}, which were published after \newcite{DBLP:journals/corr/abs-1806-00258}. 

\newcite{arthur-etal-2016-incorporating} use discrete translation lexicons to improve the performance of NMT systems for low-frequency words. Their solution involves both automatically learned lexicons, similar to those used for Statistical Machine Translation (SMT), and bilingual dictionaries, as well as a combination of both. Although their solution improves the performance of the NMT system for the English-Japanese language pair, it is not actually focused on domain adaptation, but on addressing the issue of low-frequency words, usually classified as unknown in NMT systems. As we will demonstrate in section \ref{sec:solutions}, Scylla performs domain adaptation substitutions even for frequent lexical units known by the NMT model.

\newcite{DBLP:journals/corr/ZhangZ16c} use bilingual dictionaries to generate pseudo sentence pairs that are fed into the NMT system during training. Scylla, on the other hand, does not require any dataset to be generated from the qualia-enriched FrameNet lexicon used, nor any additional training of the NMT system.

\newcite{DBLP:journals/corr/abs-1709-02184} compare the performance of SMT and NMT systems in the task of translating domain-specific terms out of any context. In the experiments they conducted with NMT models, they used a bilingual lexicon in the form of correspondence tables for substituting unknown words in OpenNMT \cite{klein-etal-2017-opennmt}. They report improvement in performance for some experimental setups, but, once again, the solution is focused exclusively on unknown words.

\newcite{moussallem2019utilizing} propose a methodology to incorporate Knowledge Graphs (KGs) into NMT models in two steps. First, the connect named entities in parallel corpora used for training the NMT system to a reference KG using a multilingual entity linking system. Then, they concatenate the KG embeddings into the embeddings of the NMT system. Authors report improved performance for BLEU, METEOR and \textsc{chr}F3. Nonetheless, their solution require training corpora to be annotated.  

\newcite{dougal-lonsdale-2020-improving} present a solution where a bilingual termbase is used for substituting one word in the translated sentence by another word or expression listed as an equivalent in the termbase. The algorithm is to some extent similar to that of the Scylla-T pipeline, in which regards the identification of the substitution points in the sentence. However, the solution in \newcite{dougal-lonsdale-2020-improving} does not rely on a semantically-structured database capable of telling the contexts where the terminology injection should take place apart from those where it should not. Moreover, their solution cannot perform either many-to-1 or many-to-many substitutions.

\section{A Qualia-Enriched FrameNet}
\label{sec:fn-br}

FrameNet \cite{baker-etal-1998-berkeley-framenet,framenet} is an implementation of the theory of Frame Semantics \cite{Fillmore1982} in which the lexicon of the English language is modeled against a network of background scenes – or frames. Each frame is composed of frame elements (FEs), which indicate the participants and props in the scene. From the early 2000's on, the framenet model has been expanded into other languages \cite{baker-lorenzi-2020-exploring}, FrameNet Brasil (FN-Br) being the Brazilian Portuguese branch of this initiative \cite{Torrent2018}. 

\begin{figure}
    \centering
    \includegraphics[width=1\columnwidth]{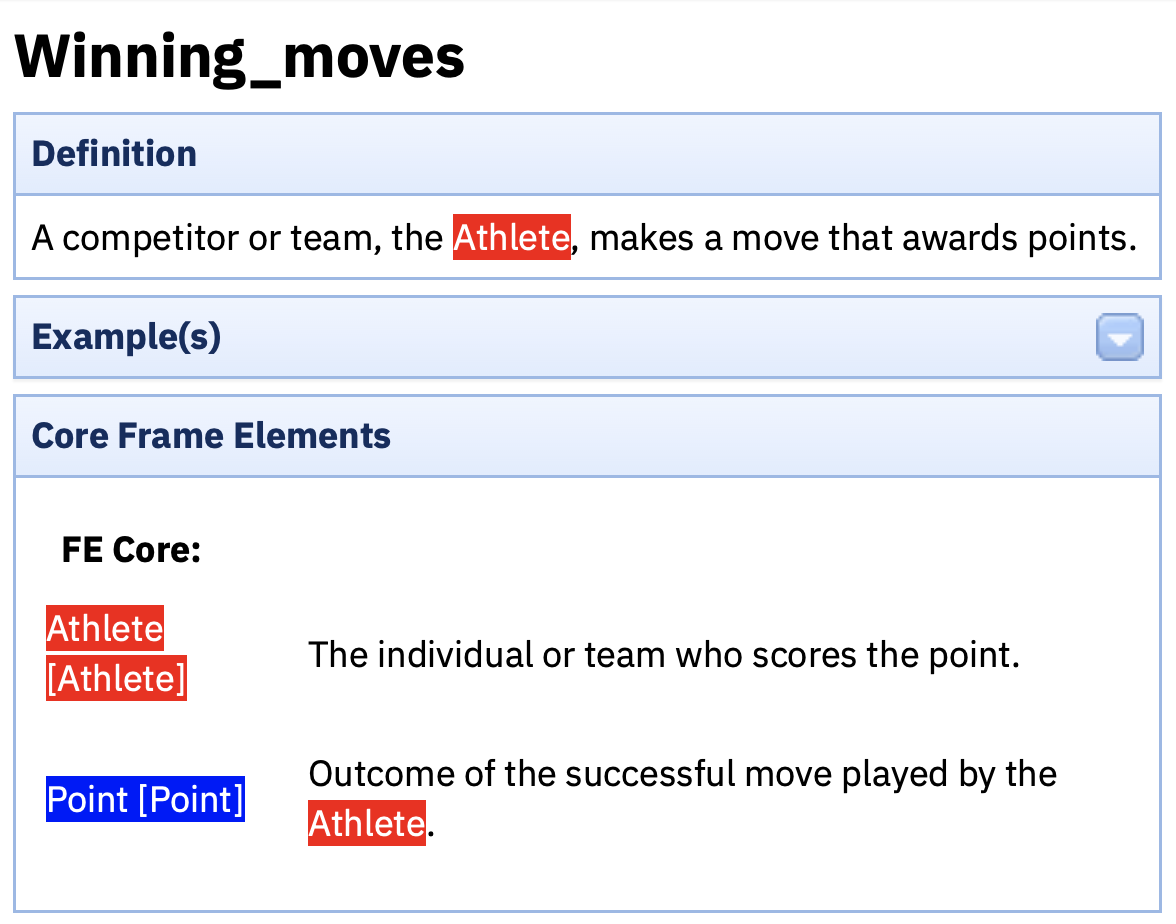}
    \caption{The \texttt{Winning\_moves} frame in FN-Br}
    \label{fig:winning}
\end{figure}

On top of expanding the framenet model into br-pt, FN-Br also develops multilingual frames for specific domains, such as Tourism and Olympic Sports \cite{torrent2014multilingual,Costa2017,Costa2018}. Because the human experience in those domains tends to abide by highly internationalized standards, frames in them also tend to be cross-linguistically applicable \cite{torrent2014multilingual}. Therefore, one same structure, such as the \texttt{Winning\_moves} frame depicted in Figure \ref{fig:winning}, can be evoked by both br-pt lexical units (LUs) like \emph{bandeja.n, gol.n} and \emph{marcar.v} and their en equivalents \emph{lay-up.n, goal.n} and \emph{score.v}, respectively.

In any framenet, frames are connected to each other via typed relations such as inheritance, subframe, perspective on, using, among others \cite{ruppenhofer2016framenet}. The \texttt{Winning\_moves} frame, for instance, inherits the \texttt{Moves} frame, since it models specific kinds of moves that result in scoring a point. In turn, \texttt{Moves} uses \texttt{Athletes}, since any LU indicating a move will to some extent make reference to the athlete performing it.  

Although Frame-to-Frame relations capture important aspects of meaning in a framenet, they are not capable of representing all the semantic relations need to properly model a domain. For instance, the original FrameNet model has no means of representing that the \textsc{Athlete} FE in the \texttt{Winning\_moves} frame can be defined in terms of the \texttt{Athletes} frame. This is to say that LUs evoking the latter, will most likely be the prototypical fillers of the \textsc{Athlete} FE in the former. FN-Br captures this information via a FE-to-Frame relation.

Moreover, relations that are specific for a set of LUs within the frames are not captured by Frame-to-Frame relations either. In the example being discussed, the relations connecting \texttt{Winning\_moves} to \texttt{Athletes} via the \texttt{Moves} frame are not able to represent that a \emph{lay up} is a winning move performed by a \emph{basketball player}, but not by a \emph{soccer player}. To address this issue, FN-Br uses qualia relations \cite{Pustejovsky1995} to connect LUs in different frames. 

The four original qualia proposed by \newcite{Pustejovsky1995} – agentive, constitutive, formal, and telic – are very general and a collection of efforts have been made to specify them \cite{lenci2000simple,pustejovsky-etal-2006-towards}. Instead of building or relying on an external ontology for refining the meaning of qualia relations, FN-Br specifies each type of quale by resorting to a frame that mediates the quale connecting two LUs. For example, according to \newcite{Pustejovsky1995}, the relation held between \emph{lay up.n} in the \texttt{Winning\_moves} frame and \emph{basketball player.n} in the \texttt{Athletes} frame would be a telic one, since scoring the point is the intention of the athlete performing the move. In the FN-Br database, those two LUs are connected via a ternary telic relation mediated by the \texttt{Intentionally\_act} frame, which features two core FEs: the \textsc{Agent} and the \textsc{Action} they perform. In the ternary quale being discussed, the LU \emph{basketball player.n} is connected to the first FE, while \emph{lay up.n} is framed by the second\footnote{The FN-Br database currently features 41 types of ternary qualia relations (TQRs), which are listed in Appendix A.}.

The qualia-enriched framenet model of the Sports domain used by Scylla was developed by \newcite{costa2020}, and features 36 frames, which can be evoked by 1,651 LUs in br-pt and 2,051 in en. A total of 4,332 instances of TQRs were created for br-pt, being then automatically replicated for en. Because some br-pt LUs may have more than one translation equivalent in en, the replication procedure yielded a total of 6,882 instances of TQRs. The distribution of TQRs per quale and per language is shown in Table \ref{table:instances}. 

\begin{table}
\centering
\begin{tabular}{lllll}
\hline
 & \textbf{Const.} & \textbf{Agent.} & \textbf{Telic} & \textbf{Formal} \\
\hline
br-pt & 950 & 82 & 1,868 & 1,432 \\
en & 1,290 & 118 & 2,462 & 3,012 \\
\hline
\end{tabular}
\caption{TQRs created for the Sports domain}{\label{table:instances}}
\end{table}

\section{Scylla: domain adaptation using frames and qualia}
\label{sec:solutions}

The methodology proposed in this paper for addressing the issue of domain adaptation in NMT systems is presented in two alternative pipelines, Scylla-S and Scylla-T, which can work around any NMT API.\footnote{For the implementations reported in this paper the NMT system used is the Google Translate V2 API.} For both of them, the most fundamental step is that of identifying the frame evoked by each LU in the sentence to be translated and, for Scylla-T, also in the translation alternatives provided by the NMT system. Next, we present how frame assignment is performed in the Scylla pipelines.

\subsection{Frame disambiguation}
\label{sec:daisy}

Using frames for MT requires identifying which frames are evoked by the lexical material in the sentence to be translated. This step was implemented in the Scylla pipelines through DAISY: \textit{Disambiguation Algorithm for Inferring the Semantics of Y}. DAISY uses the FN-Br network of typed relations as a graph, applying spread activation to estimate the frame associated with each lemma in the sentence.

Graph construction involves the following steps:
\begin{enumerate}
    \item A dependency parser – UDPipe \cite{straka-strakova-2020-udpipe} – processes the input sentence for the identification of word forms and lemmas in the sentence;
    \item From the lemmas obtained, the system searches for MWEs matching those in the FN-Br database;
    \item Based on a simplified set of syntactic patterns, lemma clusters are defined. Each cluster contains the lemmas which are directly associated to each other;
    \item LUs associated to each lemma are retrieved from the FN-Br database;
    \item Qualia relations holding between LUs in a cluster are retrieved;
    \item The frame evoked by each LU is retrieved form the FN-Br database, for each frame, related frames are also stored;
    \item FE-to-Frame relations are retrieved so that the frame evoked by the LU can be related to other frames evoked by other LUs in the cluster.
\end{enumerate}
All those elements (word forms, lemmas, LUs, and frames) are used as nodes in the graph built and the relations between them are the links connecting the nodes.

DAISY uses the spread activation search method to traverse the graph. The process is initiated by posing an “energy level” or “activation” to a set of initial nodes and then iteratively propagating that activation out to other nodes linked to the source nodes. Every time a link is traversed, the activation values decay according to a predefined formula. If a target node receives activation from more than one source node, its activation value is incremented. This method allows for measuring the relative importance of each node in the network, considering not only how much the node is distant from the initial nodes, but also how it connects to other nodes.

Spread activation for DAISY uses real activation values. The initial nodes receive an activation value of 1.0. Each link type is assigned a weight. This weight is used to decrease the activation value. Current implementation applies the following weights to relations: (i) frame evocation: 1.0; (ii) frame inheritance: 1.0; (iii) frame perspective: 0.9; (iv) subframe: 0.8; (v) frame element to frame: 0.5; (vi) qualia relations: 0.9. For each iteration \emph{p}, a given node \emph{j} has an activation level represented by \(A_{k}(p)\) and generates an output \({O_{j}(p)}\), being that a function of its activation level, according to Equation \ref{eq:spread}.

\begin{equation}
\label{eq:spread}
A_{k}(p)=\sum_{j}{O_{j}(p-1)W_{jk}}
\end{equation}

The output function \({O_{j}(p)}\), a variation of the logistic function represented in \ref{eq:output}, was carefully chosen to avoid excessive activation of the nodes in the network. 

\begin{equation}
\label{eq:output}
O_{j}(p)= \frac{1 - exp(5 * (-A_{j}(p)))}{1 + exp(-A_{j}(p))}
\end{equation}

The spread activation process occurs until it reaches the frames. At this point, a backpropagation process is initiated, applying the activation level calculation again at each node until reaching the LUs. Hence, each LU is assigned a weight indicating its relative importance in the network. Finally, the relative weight of each LU is associated to each lemma. This is made by adding up the activation level of each LU and dividing the result by the number of related LUs. The frame evoked by the LU with the highest relative weight is taken as the frame associated with the lemma. 

The frame assignment and/or disambiguation process performed by DAISY has the advantage of not depending on large annotated datasets for training, as it is the case of SEMAFOR \cite{chen-etal-2010-semafor} and Open Sesame \cite{swayamdipta2017frame}. Moreover, because it takes the sentence context into consideration when assigning the best fit frame for polysemous lemmas, it helps avoid performance loss for out of domain sentences. Consider, for example, the br-pt sentences in \ref{ex:bandejabasquete} and \ref{ex:bandejanormal}. Note that \emph{bandeja.n} is a polysemous lemma in br-pt. While in \ref{ex:bandejabasquete} it evokes the \texttt{Winning\_moves} frame, in \ref{ex:bandejanormal}, it evokes \texttt{Utensils}. DAISY is capable of correctly identifying each of those frames because of the surrounding context in each sentence.\footnote{The frame assignment graphs generated by DAISY for sentences in \ref{ex:bandejabasquete} and \ref{ex:bandejanormal} are given in Appendix B.} 

\ex.O jogador de basquete converteu a \textbf{bandeja}. \\ \emph{The basketball player scored the lay-up.}
\label{ex:bandejabasquete}

\ex.O garçom colocou as tijelas na \textbf{bandeja}. \\ \emph{The waiter put the bowls on the tray.}
\label{ex:bandejanormal}

Both versions of Scylla use DAISY to acquire information on the frames evoked by the source sentence and, for the Scylla-T version, also by the target sentence. Next, we present each version of the pipeline in detail.

\subsection{Scylla-S: terminology injection during the pre-processing stage}
\label{sec:pre}

In Scylla-S, the process of terminology injection occurs in a pre-processing stage. The source sentence is submitted to DAISY and Scylla-S substitutes the source words or MWEs for which it found an entry in the FN-Br database for their translation equivalents. The hybrid sentence is then submitted to the NMT system, which, in turn, is set so that it copies unknown words into the target sentence. Because most translations are not part of the set of known words in the source language, the resulting sentence usually contains the domain-adequate expression. 

\begin{figure}
    \centering
    \includegraphics[width=1\columnwidth]{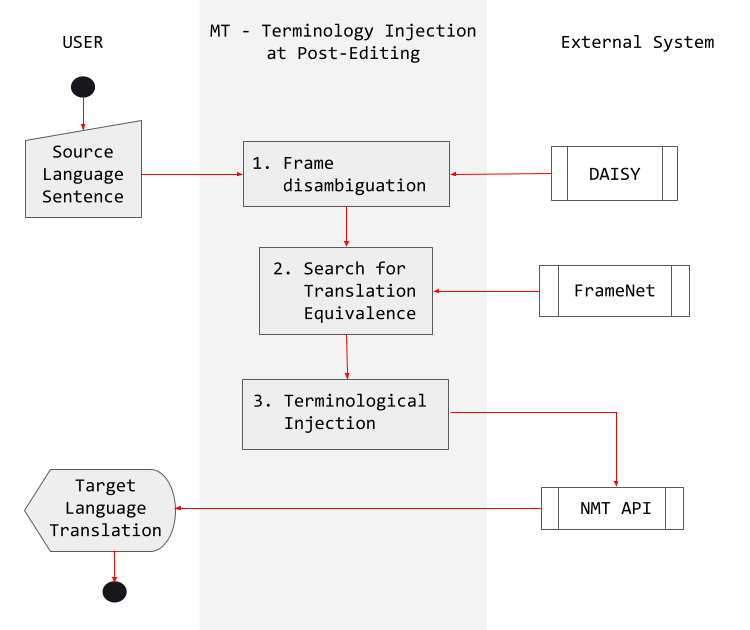}
    \caption{The Scylla-S pipeline}
    \label{fig:scyllas}
\end{figure}

For example, when the source sentence in \ref{ex:pontasource} is submitted to Scylla-S, it generates the hybrid sentence in \ref{ex:pontahybrid}. This hybrid sentence is then submitted to the NMT API, yielding \ref{ex:pontascyllas} as an output. A summary of Scylla-S is presented in Figure \ref{fig:scyllas}.

\ex.O ponta é o jogador que menos tempo tem para pensar na armação de uma jogada. \\
\emph{The wing is the player with less time think about setting up a play}
\label{ex:pontasource}

\ex.O \textbf{wing} é o \textbf{player} que menos tempo tem para pensar na armação de uma \textbf{play}.
\label{ex:pontahybrid}

\ex.The wing is the player \textbf{that has less time to think in the setup of} a play.
\label{ex:pontascyllas}

Scylla-S has some limitations. First, because the sentence fed into the NMT system is a hybrid – \ref{ex:pontahybrid} –, containing words in both the source and the target languages, the performance of the NMT system decays sensibly, mainly regarding fluency – see boldface fragment in \ref{ex:pontascyllas}. Moreover, sometimes the target language word happens to be found in the source language vocabulary of the NMT system. For instance, one of the sentences in the experimental dataset had the LU \emph{levantamento.n}, which, in the domain of weightlifting, is equivalent to \emph{lift.n}. However, when the hybrid sentence is fed to the NMT API, it recognizes \emph{lift.n} as a br-pt word used in the domain of plastic surgery and, therefore, translates it again into \emph{facelift.n}.

In an attempt to overcome the limitations of Scylla-S, we developed Scylla-T.

\begin{figure}
    \centering
    \includegraphics[width=1\columnwidth]{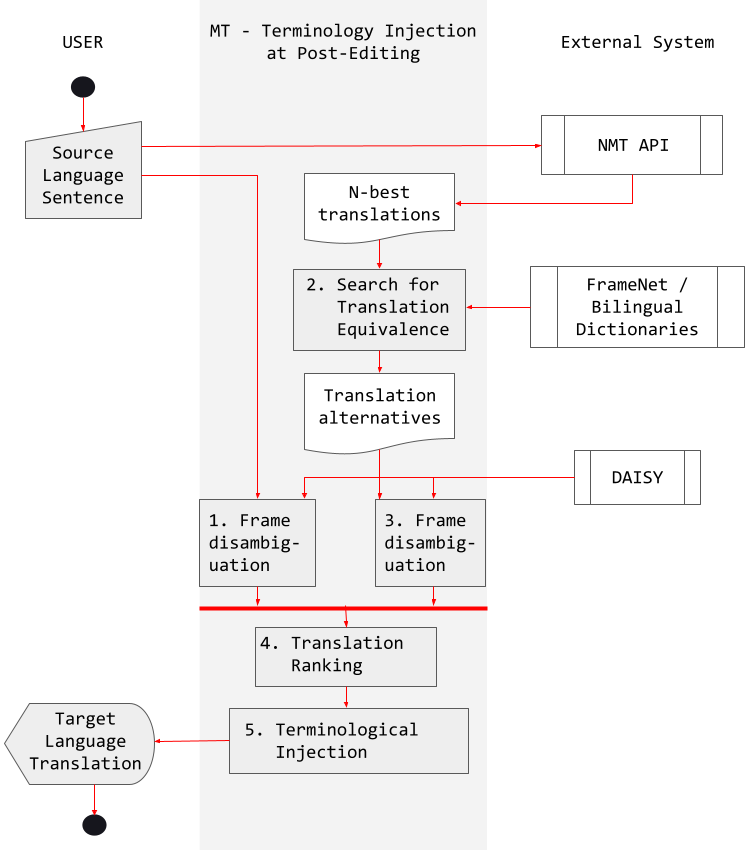}
    \caption{The Scylla-T pipeline}
    \label{fig:scyllat}
\end{figure}

\subsection{Scylla-T: terminology injection during the post-editing stage}
\label{sec:post}

Scylla-T performs terminology injection in the target sentence. For this process to work properly, it is necessary to align the words and MWEs in the source sentence with those in the sentence generated by the NMT API. First, the source sentence is submitted to the NMT API and n-best translations are retrieved. Next, Scylla-T queries the FN-Br database and also a bilingual dictionary API\footnote{For the implementation reported in this paper, the Oxford Dictionary API was used under a free academic license.} for all possible translation equivalents of the words and MWEs in the n-best translations generated by the NMT API. The bilingual dictionary is used as means of guaranteeing that words not included in the FN-Br database are also covered by the process. 

The translation equivalents retrieved are compared to the original words or MWEs in the source sentence and their synonyms in the source language. Whenever the system finds a match, an alignment is created for a word pair. The Jaro-Winkler similarity metric is used so that different word forms are considered when generating the matches.\footnote{Jaro-Winkler was preferred over Levenshtein because the first assigns higher values to word pairs that are more similar towards the left frontier. Because we wanted to preserve the performance of the NMT API in correctly inflecting words in the target language, Jaro-Winkler was chosen.} Once the alignment is concluded, equivalence sets are concatenated to compose the set of translation alternatives for each word in the target sentence.

Such an approach is based on a binary tree search, where each node represents a possible point for terminology injection at the target sentence. When building the tree, nodes are created for each translation alternative. To avoid node duplicity and to assure that no translation alternative is ignored, tree construction is performed recursively, so that every possible combination of terminology injection for one same word or MWE is considered.

Because the binary tree search aims to find the translation alternative with the highest semantic similarity with the source sentence, the definition of an objective function is necessary. We defined a semantic similarity metric based on the frames evoked by each sentence, which were extracted using DAISY. Equation \ref{eq:sem_sim} computes the semantic similarity between two sentences from the number of coincidental frames found in both. $F_{o}$ and $F_{d}$ are the set of frames extracted by DAISY for the source and target sentences, respectively. The binary tree search results in the maximization of Equation \ref{eq:sem_sim} and is used as a heuristics for terminology injection in the target sentence, either by re-ranking one of the n-best sentence translation alternatives, or by the substitution of equivalents that are not domain-compliant by in-domain terms form the FN-Br database. Therefore, the final output of Scylla-T is a translation with a higher semantic similarity with the source sentence.

\begin{equation}
\label{eq:sem_sim}
    S(F_{o}, F_{d}) = \sum_{f_{o} \in F_{o}}^{|F_{o}|}\sum_{f_{d} \in F_{d}}^{|F_{d}|} 1 [f_{o} = f_{d}]
\end{equation}

Scylla-T improves the performance of Scylla-S in three ways. First, because it avoids the submission of hybrid sentences to the NMT API. Second, because it may operate by just re-ranking one of the n-best sentence translation alternatives generated by the NMT, preserving its performance in which concerns fluency. Third, because, if it needs to substitute a term by another one that is domain-compliant, it does it in a more localized and precise fashion, avoiding performance loss caused by terminology injection. Those improvements can be exemplified in the translation generated by Scylla-T to the source sentence in \ref{ex:pontasource}, given in \ref{ex:pontascyllat}. When compared to the translation generated by the NMT API alone, given in \ref{ex:pontanmt}, \ref{ex:pontascyllat} is equally fluent, but terminologically accurate, since \emph{ponta.n} in br-pt is to be translated as \emph{wing.n} or \emph{winger.n} in en, not as \emph{forward.n}.

\ex.The winger is the player who has less time \textbf{to think about setting up} a play.
\label{ex:pontascyllat}

\ex.The forward is the player who has less time to think about setting up a move.
\label{ex:pontanmt}


\section{Evaluation}
\label{sec:evaluation}

To evaluate the performance of Scylla-S and Scylla-T, a sentence translation experiment for the br-pt–en language pair was designed. Both pipelines were evaluated against a commercial NMT API, namely the Google Translate V2 API, used as a baseline for BLEU, TER, and HTER. Dataset, experimental setup and results are presented next.

\subsection{Dataset}
\label{sec:dataset}

The dataset used was specifically assembled for the experiment reported in this paper. It is composed by a set of sentences of the Sports domain in br-pt and a reference translation for each of them in en.\footnote{All sentences in the dataset are available at https://github.com/FrameNetBrasil/scylla\_lr. See also section \ref{sec:eth}.}  

\paragraph{Source language sentences:}
For conducting the translation experiment, a dataset containing 50 br-pt sentences from the Sports domain was created. All sentences were extracted from naturally produced texts, that is, from instances of textual genres such as news, encyclopedias, blog posts etc. produced by native speakers of br-pt and published in newspapers, magazines, books, websites, and the like. To make sure that the performance of the systems evaluated for domain adaptation is actually measured, for each sentence, there was at least one polysemous lemma, with at least two possible meanings, one of which related to the Sports domain. The average number of polysemous lemmas per sentence was 2.16, and the average number of possible meanings per polysemous lemma was 2.17.

\paragraph{Reference translation:}
The 50 br-pt sentences were then translated into English by a professional translator who is a native speaker of English. The translated sentences were then revised for morphosyntactic aspects and fluency by four native speakers of English. Next, sentences were analyzed for frame preservation in the Sports domain, following the Primacy of the Frame model of translation \cite{czulo2017aspects}. This is to say that, for each sentence, one linguist checked whether the frames evoked by the source and the target sentences were the same, regarding the Sports domain. The percentage of frame preservation was 72.4\% of all Sports frames evoked. Provided that it is expected that translations may reorganize sentence structure, suppressing words or substituting them by hypernyms, for example, such a percentage is indicative of high semantic similarity. The reference translation sentences were used as a gold standard for evaluating the performance of the proposed solutions, as described next. 

\subsection{Experiments}
\label{sec:experiments}

To evaluate the performance of Scylla-S and Scylla-T, the 50 sentences in br-pt were submitted to the pipelines described in \ref{sec:pre} and \ref{sec:post}. Sentences are also submitted to NMT API used in Scylla-S and Scylla-T, which is considered as the baseline for comparison. Experiments were performed on a Ubuntu 20 system with 20GB RAM and a 3.1 GHz Intel i5 7200U processor.\footnote{The baseline system used can be accessed at https://cloud.google.com/translate.}

Performance was evaluated for BLEU \cite{papineni-etal-2002-bleu}, TER, and HTER \cite{snover-etal-2006-study}. BLEU evaluates correspondences between n-grams in the translations produced by each system and the gold standard translations. According to the interpretation criteria for this metric, the higher the score, the better the translation. This metric does not focus on evaluating the preservation of the semantics of the source sentence in the target sentence. Because our aim is to evaluate the performance of Scylla-S and Scylla-T for domain adaptation in NMT, we chose two other metrics – TER and HTER – that measure the effort required in post-editing a sentence generated by a MT system so that it can be regarded as a fluent, adequate translation of the source sentence. 

TER uses the gold standard translations as reference to compute the minimal number of edits – additions, deletions, substitutions – that would be required in the sentence produced by the MT system so that it matches the reference sentence. Because TER is unable to assess the semantics of the MT sentence in contrast with that of the gold standard, it may end up proposing edits that are not needed, since the MT sentence just presents a different, but yet fluent and adequate, translation of the source sentence. 

To overcome this limitation, HTER uses expert human translators to perform the edits so that the MT sentence becomes fluent and semantically similar to the gold standard. To avoid a high influence of human subjective choices while editing the MT sentences taking the gold standard sentences as a reference, three professional translators were hired for the task. Each of the three professional translators independently edited each of the 150 machine translated sentences (50 from each system used) taking the gold standard translations as a reference. The calculation of HTER was then based on the average number of edits proposed for each sentence, given the number of edits made by each professional translator. The computation of the number and types of edits was also carried out independently by two reviewers and revised by a third person.

Since TER and HTER measure the effort required for editing the MT sentence, the lower the score, the better the translation.

\subsection{Results and Discussion}
\label{sec:results}

The performance of the three systems evaluated in the experiment for the three metrics chosen is presented in Table \ref{table:evaluation}.

\begin{table}
\centering
\begin{tabular}{lrrr}
\hline
 & \textbf{Baseline} & \textbf{Scylla-S} & \textbf{Scylla-T} \\
\hline
BLEU & 53.13 & 48.12 & \textbf{53.66} \\
TER & \textbf{36.23} & 42.63 & 36.47 \\
HTER & 13.80 & 10.44 & \textbf{7.38} \\
\hline
\end{tabular}
\caption{Evaluation of the baseline, Scylla-S and Scylla-T systems for BLUE, TER and HTER}{\label{table:evaluation}}
\end{table}

The performance of Scylla-S is the worst among the three systems for BLEU. This may be due to the fact that Scylla-S feeds hybrid sentences to the NMT API, compromising its performance at the formal pole of language structures. Even considering that BLEU does not evaluate semantic adequacy directly, the Baseline and Scylla-T have a similar performance for this metric. 

The morphosyntactic problems caused by sentence hybridization in Scylla-S are also the reason for a poorer performance of this system when evaluated for TER. Because TER is applied automatically and computes all edits needed to turn the MT output into an exact copy of the reference translation, the gains that derive from the fact that sentences produced by Scylla-S are terminologically adequate to the domain are surpassed by the losses in fluency. Once again, the Baseline and Scylla-T have a superior and very similar performance for this metric. This demonstrates that Scylla-T is able to mitigate the problems caused by the fact that Scylla-S feeds hybrid sentences to the NMT API.

For HTER, both Scylla-S and Scylla-T outperform the Baseline system. This is due to the fact that, when professional human translators are asked to adequate the MT sentence using the domain-specific gold standard as a reference, they take into consideration the fact that polysemous lemmas in the source language may have different translations in the target language, depending on the domain. Because sentences were extracted from in-domain real texts, the context provided by the sentences was able to indicate that the translation should be adequate to the Sports domain. Moreover, human translators are able to consider a sentence containing, for instance, terms that are synonymous to the ones in gold standard as equally adequate to the domain. Therefore, the performance of the systems measured by HTER is considerably better than that measured by TER.

Given the nature of the problem tackled in this paper, that of domain adaptation in NMT, we claim that HTER, although having a higher cost, is the most reliable metric for assessing the performance of the three systems in the task proposed. To illustrate, consider once again the examples already discussed in sections \ref{sec:pre} and \ref{sec:post}, presented together with the gold standard translation in (8-12). 

\ex.O ponta é o jogador que menos tempo tem para pensar na armação de uma jogada. \\
\emph{Source sentence}

\ex.The winger is the player with less time to think about setting up a strike. \\ 
\emph{Gold standard translation}

\ex.The \textbf{forward} is the player who has less time to think about setting up a move. \\ 
\emph{Baseline (TER=26.66 / HTER=0.08)}

\ex.The wing is the player that has less time \textbf{to think in the setup of} a play. \\
\emph{Scylla-S (TER=53.33 / HTER=0.06)}

\ex.The winger is the player who has less time to think about setting up a play. \\
\emph{Scylla-T (TER=20.00 TER / HTER=0.00)}

Note that the translation generated by the Baseline System does not feature an adequate translation for \emph{ponta.n} in the domain, while the ones generated by Scylla-S and Scylla-T do. However, the translation by Scylla-S is less fluent, while the one by Scylla-T is not, leading to lower TER and HTER values.

From the performance results for HTER in Table \ref{table:evaluation}, we conclude that the domain adapted translations generated by both Scylla-S and Scylla-T present better semantic correlation with the gold standard. Scylla-T was also able to address the limitations of Scylla-S, improving the performance of the Baseline NMT system by almost 47\%. 

\section{Conclusions and Outlook}
\label{sec:conclusions}

In this paper, we presented two systems for domain adaptation in NMT using terminology injection based on qualia-enriched FrameNet. Neither system require fine-tuning of the NMT model, reducing computational costs, and mitigating performance losses due to over-fitting. Evaluation of the systems' performance against a NMT API taken as baseline for the BLEU, TER, and HTER metrics demonstrated that one of the systems, Scylla-T, meets the performance of the baseline for both BLEU and TER. For the HTER metric, more adequate for measuring semantic adequacy, both systems proposed outperform the baseline, Scylla-T improving it by 47\%.

Scylla-S and Scylla-T are the first systems to our knowledge to leverage framenet data and qualia relations for domain adaptation in NMT. Also, because they are both implemented as pipelines, they can be easily adapted to any NMT API.

For future work, we plan to explore the automatic expansion of domain-specific framenet coverage for other languages via the acquistion of LUs for languages still not covered by FN-Br from other lexical resources such as multilingual WordNets \cite{fellbaum2012challenges}, or BabelNet \cite{navigli2021ten}. The idea is that, because the FN-Br database already models equivalences between domain-specific lexical items in br-pt and en, those equivalences could be used as proxies for extracting equivalent terminology in other languages, making Scylla available in other languages without the need to rebuild the domain model from scratch.    

\section{Ethics Statement}
\label{sec:eth}

Both the gold standard translations and the evaluation based on the HTER metric were carried out by professional translators working for a translation company hired for that purpose. Translators worked under standard working contract regulations. Experiments were conducted without any major computational costs, as it can be inferred from the specifications of the computational environment provided in \ref{sec:experiments}.

\section{Acknowledgements}

The FrameNet Brasil lab is funded by CAPES PROBRAL grant 88887.144043/2017-00. Diniz da Costa's research was funded by CAPES PROBRAL PhD exchange grant 88887.185051/2018-00. Authors thank Oliver Czulo and Alexander Ziem for their contribution to the development of this research project.

\section{Bibliographical References}\label{reference}

\bibliographystyle{lrec2022-bib}
\bibliography{scylla,anthology}

\section*{Appendices}

\subsection*{A – Ternary Qualia Relations in the FN-Br Database}
\label{sec:tqr_list}

The distribution of ternary qualia relations (TQRs) per quale and per language is shown in Table \ref{table:instances}.

Table \ref{table:tqr} presents the 41 TQRs modeled in FN-Br. The first column shows the quale refined by the TQR, while the second gives a mnemonic key for the relation. The third column shows the frame mediating the TQR, while the last two show the core FEs which can be prototypically instantiated by the two LUs involved in the TQR. 

\begin{table*}
\begin{center}
\begin{tabular}{|l|l|l|l|l|}
\hline \bf Quale & \bf Relation & \bf Frame & \bf FE1 & \bf FE2 \\ \hline
Agentive & created by & \texttt{Achieving\_first} & \textsc{New\_idea} & \textsc{Cognizer} \\
Agentive & caused by & \texttt{Causation} & \textsc{Effect} & Actor \\
Agentive & caused by & \texttt{Causation} & \textsc{Effect} & Cause \\
Agentive & created by & \texttt{Cooking\_creation} & \textsc{Produced\_food} & \textsc{Cook} \\
Agentive & caused by & \texttt{Intentionally\_act} & \textsc{Act} & \textsc{Agent} \\
Agentive & affected by & \texttt{Intentionally\_affect} & \textsc{Patient} & \textsc{Agent} \\
Agentive & created by & \texttt{Intentionally\_create} & \textsc{Created\_ent.} & \textsc{Creator} \\
Agentive & resolved by & \texttt{Resolve\_problem} & \textsc{Problem} & \textsc{Agent} \\
Constitutive & has as attribute & \texttt{Attributes} & \textsc{Entity} & \textsc{Attribute} \\
Constitutive & has as part & \texttt{Building\_parts} & \textsc{Whole} & \textsc{Part} \\
Constitutive & causes & \texttt{Causation} & \textsc{Actor} & \textsc{Affected} \\
Constitutive & contains & \texttt{Containing} & \textsc{Container} & \textsc{Contents} \\
Constitutive & produces & \texttt{Creating} & \textsc{Creator} & \textsc{Created\_ent.} \\
Constitutive & workplace of & \texttt{Employing} & \textsc{Employer} & \textsc{Employee} \\
Constitutive & includes & \texttt{Inclusion} & \textsc{Total} & \textsc{Part} \\
Constitutive & used by & \texttt{Infrastructure} & \textsc{Infrastructure} & \textsc{User} \\
Constitutive & made of & \texttt{Ingredients} & \textsc{Product} & \textsc{Material} \\
Constitutive & performed by & \texttt{Intentionally\_act} & \textsc{Act} & \textsc{Agent} \\
Constitutive & relative of & \texttt{Kinship} & \textsc{Ego} & \textsc{Alter} \\
Constitutive & has as member & \texttt{Membership} & \textsc{Group} & \textsc{Member} \\
Constitutive & affects & \texttt{Obj\_influence} & \textsc{Influencing\_ent.} & \textsc{Dependent\_ent.} \\
Constitutive & has as part & \texttt{Part\_inner\_outer} & \textsc{Whole} & \textsc{Part} \\
Constitutive & has as part & \texttt{Part\_piece} & \textsc{Substance} & \textsc{Piece} \\
Constitutive & has as part & \texttt{Part\_whole} & \textsc{Whole} & \textsc{Part} \\
Constitutive & has origin at & \texttt{People\_origin} & \textsc{Person} & \textsc{Origin} \\
Constitutive & follower of & \texttt{People\_religion} & \textsc{Person} & \textsc{Religion} \\
Constitutive & relates to & \texttt{Relation} & \textsc{Entity\_1} & \textsc{Entity\_2} \\
Constitutive & has as resident & \texttt{Residence} & \textsc{Location} & \textsc{Resident} \\
Constitutive & uses & \texttt{Using\_resource} & \textsc{Agent} & \textsc{Resource} \\
Formal & instance of & \texttt{Exemplar} & \textsc{Instance} & \textsc{Type} \\
Formal & type of & \texttt{Type} & \textsc{Subtype} & \textsc{Category} \\
Telic & vice of & \texttt{Addiction} & \textsc{Addictant} & \textsc{Addict} \\
Telic & ability of & \texttt{Capability} & \textsc{Event} & \textsc{Entity} \\
Telic & habit of & \texttt{Custom} & \textsc{Behavior} & \textsc{Protagonist} \\
Telic & performed at & \texttt{Infrastructure} & \textsc{Activity} & \textsc{Infrastructure} \\
Telic & activity of & \texttt{Intentionally\_act} & \textsc{Act} & \textsc{Agent} \\
Telic & created by & \texttt{Intentionally\_create} & \textsc{Created\_ent.} & \textsc{Creator} \\
Telic & purpose of & \texttt{Purpose} & \textsc{Goal} & \textsc{Agent} \\
Telic & purpose of & \texttt{Tool\_purpose} & \textsc{Purpose} & \textsc{Tool} \\
Telic & used for & \texttt{Using} & \textsc{Agent} & \textsc{Purpose} \\
Telic & used by & \texttt{Using\_resource} & \textsc{Resource} & \textsc{Agent} \\
\hline
\end{tabular}
\end{center}
\caption{\label{table:tqr} Ternary Qualia relations in FN-Br. }
\end{table*}

\subsection*{B – Example frame assignment graphs generated by DAISY.}
\label{sec:exgraphs}

\begin{figure*}
    \centering
    \includegraphics[width=1\textwidth]{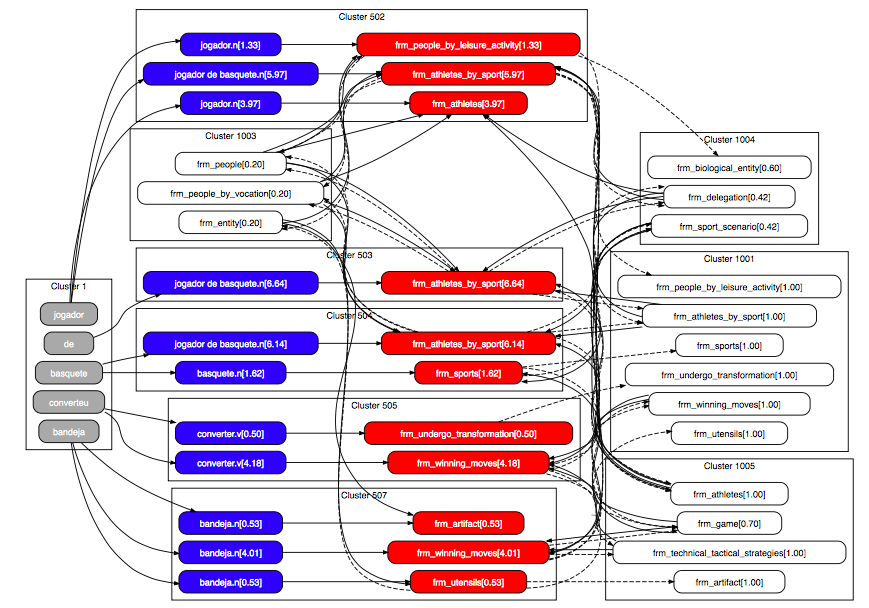}
    \caption{The frame assignment graph generated by DAISY for example sentence \ref{ex:bandejabasquete}.}
    \label{fig:daisylayup}
\end{figure*}

\begin{figure*}
    \centering
    \includegraphics[width=1\textwidth]{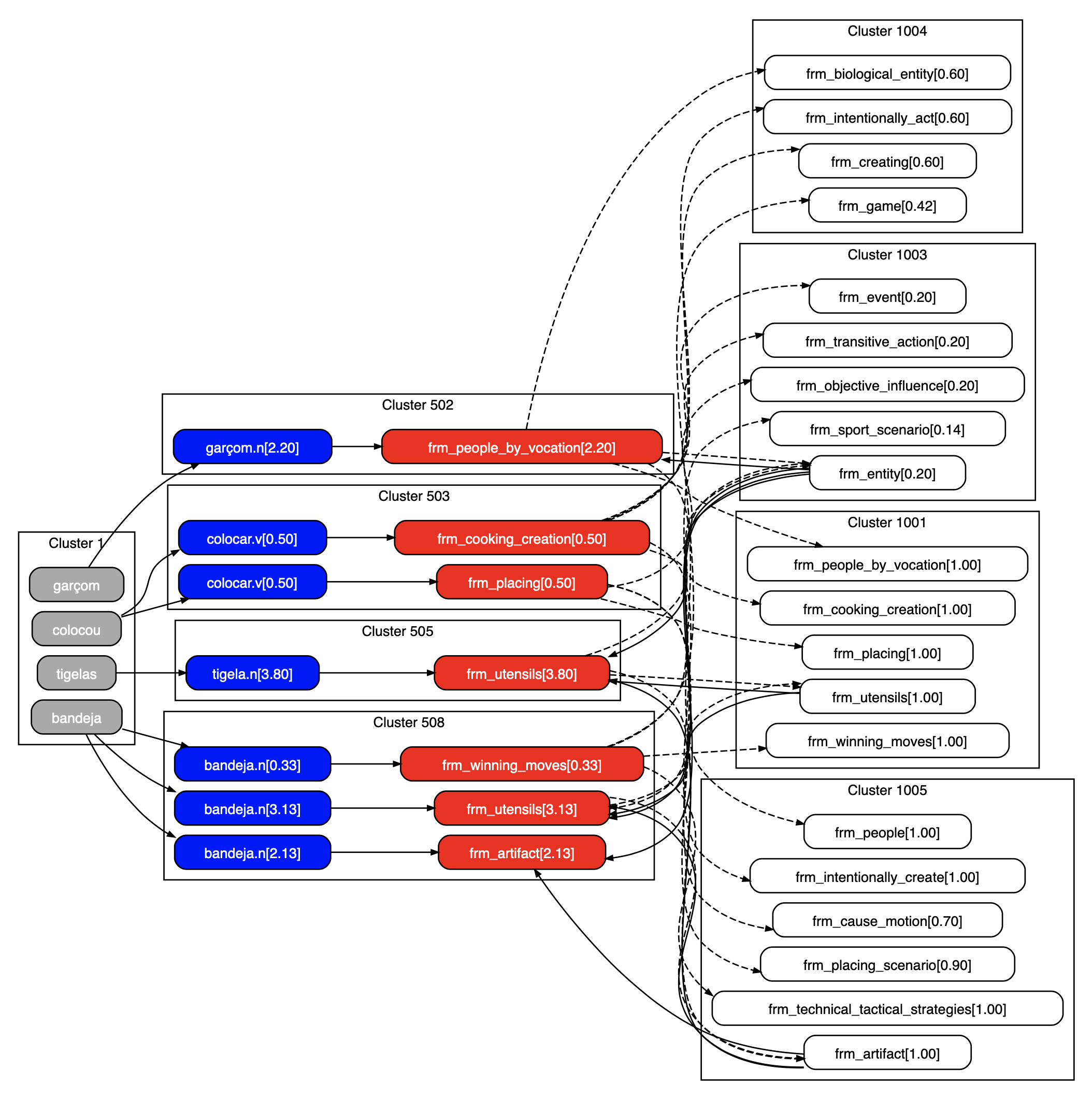}
    \caption{The frame assignment graph generated by DAISY for example sentence \ref{ex:bandejanormal}.}
    \label{fig:daisytray}
\end{figure*}

Figures \ref{fig:daisylayup} and \ref{fig:daisytray} depict the graphs generated by DAISY for assigning the correct frames to \emph{bandeja.n}, according to the context provided by each sentence. Both figures were generated using the DAISY demo interface available at http://server3.framenetbr.ufjf.br:8010/index.php/ daisy/main.

Note, in Figure \ref{fig:daisylayup}, that DAISY is able to properly disambiguate the lemma \emph{bandeja.n}, shown in cluster 507. In the FN-Br lexicon, this lemma may evoke three different frames: \texttt{Artifact} and \texttt{Utensils}, being equivalent to \emph{tray.n} and \texttt{Winning\_moves}, being equivalent to \emph{lay up.n}. In the context of a sentence where the subject is a basketball player, the last frame is the correct one. DAISY reaches an activation level of 4.01 for the \texttt{Winning\_moves} frame, against a 0.53 level for the other two. DAISY also correctly disambiguates between the two senses of \emph{converter.v} – cluster 505 –, assigning an activation level of 4.18 to the \texttt{Winning\_moves} frame, against only 0.50 to the \texttt{Undergo\_transformation} frame. In en, the equivalent LUs evoking those frames would be \emph{score.v} and \emph{turn into.v}, respectively. Finally, it is worth noting that DAISY correctly identifies the MWE \emph{jogador de basquete.n}, equivalent to \emph{basketball player.n}, as shown in clusters 502, 503 and 504. In the first, the MWE is preferred over \emph{jogador.n} alone. In the third, over \emph{basquete.n} alone.

A similar scenario holds for Figure \ref{fig:daisytray}, where DAISY correctly identifies \emph{bandeja.n} as an LU evoking the \texttt{Utensils} frame with ans activation level of 3.13.

\end{document}